\documentclass{article} 
\usepackage{iclr2021_conference,times}

\usepackage[utf8]{inputenc} 
\usepackage[T1]{fontenc}    
\usepackage{hyperref}       
\usepackage{url}            
\usepackage{booktabs}       
\usepackage{amsfonts}       
\usepackage{nicefrac}       
\usepackage{microtype}      
\usepackage{enumitem}
\usepackage{graphicx}
\usepackage{makecell}
\usepackage{algorithm}
\usepackage{algorithmic}
\usepackage{amsthm}
\usepackage{caption}
\usepackage{mathtools}

\DeclareCaptionLabelFormat{andtable}{#1~#2  \&  \tablename~\thetable}

\usepackage{amsmath,amsfonts,bm}
\allowdisplaybreaks
\newcommand\Tstrut{\rule{0pt}{2.3ex}}
\newcommand\Bstrut{\rule[-0.9ex]{0pt}{0pt}}

\newtheorem{prop}{Proposition}

\newcommand{\loss}{\mathcal{L}}

\newcommand{\set}[1]{\{ #1 \}}

\def\eqref#1{equation~\ref{#1}}

\def\1{\bm{1}}

\DeclareMathAlphabet{\mathsfit}{\encodingdefault}{\sfdefault}{m}{sl}
\SetMathAlphabet{\mathsfit}{bold}{\encodingdefault}{\sfdefault}{bx}{n}

\def\gD{{\mathcal{D}}}
\def\gE{{\mathcal{E}}}
\def\gF{{\mathcal{F}}}

\def\gP{{\mathcal{P}}}

\def\gR{{\mathcal{R}}}

\DeclareMathOperator*{\argmin}{arg\,min}

\title{Enforcing Predictive Invariance across Structured Biomedical Domains}

\author{Wengong Jin \quad Regina Barzilay \quad Tommi Jaakkola \\
CSAIL, Massachusetts Institute of Technology \\
\texttt{\{wengong,regina,tommi\}@csail.mit.edu}
}

\iclrfinalcopy 
\begin{document}
\maketitle

\begin{abstract}
    
Many biochemical applications such as molecular property prediction require models to generalize beyond their training domains (environments). Moreover, natural environments in these tasks are structured, defined by complex descriptors such as molecular scaffolds or protein families. Therefore, most environments are either never seen during training, or contain only a single training example. To address these challenges, we propose a new regret minimization (RGM) algorithm and its extension for structured environments. RGM builds from invariant risk minimization (IRM) by recasting simultaneous optimality condition in terms of predictive regret, finding a representation that enables the predictor to compete against an oracle with hindsight access to held-out environments. The structured extension adaptively highlights variation due to complex environments via specialized domain perturbations. We evaluate our method on multiple applications: molecular property prediction, protein homology and stability prediction and show that RGM significantly outperforms previous state-of-the-art baselines.

\end{abstract}

\section{Introduction}

In many biomedical applications, training data is necessarily limited or otherwise heterogeneous. It is therefore important to ensure that model predictions derived from such data generalize substantially beyond where the training samples lie. For instance, in molecule property prediction \citep{wu2018moleculenet}, models are often evaluated under scaffold split, which introduces structural separation between the chemical spaces of training and test compounds. In protein homology detection~\citep{tape2019}, the
split is driven by protein superfamily where entire evolutionary groups are held out from the training set, forcing models to generalize across larger evolutionary gaps.

\begin{figure}[t]
    \centering
    \includegraphics[width=\textwidth]{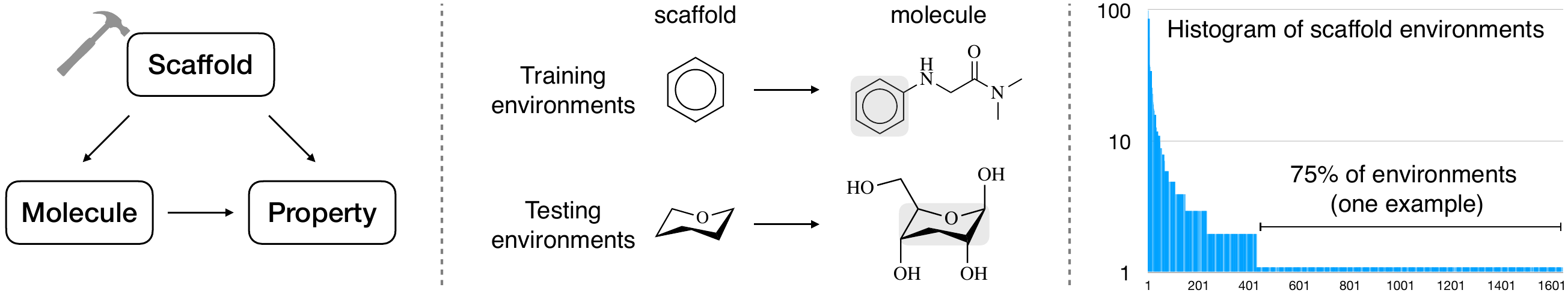}
    \vspace{-12pt}
    \caption{\emph{Left}: Data generation process for molecule property prediction. Training and test environments are generated by controlling the scaffold variable. \emph{Middle}: Scaffold is a subgraph of a molecular graph with its side chains removed. \emph{Right}: In a toxicity prediction task \citep{wu2018moleculenet}, there are 1600 scaffold environments with 75\% of them having a single example.}
    \label{fig:motivation}
\end{figure}

The key technical challenge is to be able to estimate models that can generalize beyond their training data. The ability to generalize implies a notion of invariance to the differences between the available training data and where predictions are sought. A recently proposed approach known as invariant risk minimization (IRM)~\citep{arjovsky2019invariant} seeks to find predictors that are simultaneously optimal across different such scenarios (called environments). Indeed, one can apply IRM with environments corresponding to molecules sharing the same scaffold~\citep{bemis1996properties} or proteins from the same family~\citep{pfam} (see Figure~\ref{fig:motivation}). However, this is challenging since, for example, scaffolds are structured objects and can often uniquely identify each example in the training set. 
It is not helpful to create single-example environments as the model would see any variation from one example to another as scaffold variation.

In this paper, we propose a \emph{regret minimization} algorithm to handle both standard and structured environments. The basic idea is to simulate unseen environments by using part of the training set as held-out environments $E_e$. 
We quantify generalization in terms of regret --- the difference between the losses of two auxiliary predictors trained with and without examples in $E_e$.
This imposes a stronger constraint on $\phi$ and avoids some undesired representations admitted by IRM.
For the structured environments like molecular scaffolds, we simulate unseen environments by perturbing the representation $\phi$. The perturbation is defined as the gradient of an auxiliary scaffold classifier with respect to $\phi$. The difference between the original and perturbed representation highlights the scaffold variation to the model. Its associated regret measures how well a predictor trained without perturbation generalizes to the perturbed examples. The goal is to characterize the scaffold variation without explicitly creating an environment for every possible scaffold.

Our methods are evaluated on real-world datasets such as molecule property prediction and protein classification. We compare our model against multiple baselines including IRM, MLDG~\citep{li2018learning} and CrossGrad~\citep{shankar2018generalizing}. On the QM9 dataset~\citep{ramakrishnan2014quantum}, we outperform the best baseline by a wide margin across multiple properties (41.7 v.s 52.3 average MAE) under an extrapolation evaluation. On a protein stability dataset~\citep{rocklin2017}, we achieve new state-of-the-art results compared to \citet{tape2019} (0.79 v.s. 0.73 spearman's $\rho$).

\section{Related work}
\label{sec:relwork}

\textbf{Generalization challenges in biomedical applications } The challenges of generalization have been extensively documented in this area. For instance, \citet{yang2019analyzing,tape2019,hou2018deepsf} have demonstrated that state-of-the-art models exhibit significant drop in performance when tested under scaffold or protein family split. De facto, the scaffold split and its variants~\citep{feinberg2018potentialnet} are used so commonly in cheminformatics as they emulate temporal evaluation adopted in pharmaceutical industry. Therefore, the ability to generalize to new scaffold or protein family environments is the key for practical usage of these models. Moreover,  input objects in these domains are typically structured --- e.g., molecules are represented by graphs~\citep{duvenaud2015convolutional,dai2016discriminative,gilmer2017neural}. This characteristic introduces unique challenges with respect to the environment definition for IRM style algorithms.

\textbf{Invariance } Prior work has sought generalization by enforcing an appropriate invariance constraint over learned representations. For instance, domain adversarial network (DANN)~\citep{ganin2016domain,zhao2018adversarial} enforces the latent representation $Z = \phi(X)$ to have the same distribution across different environments $E$ (i.e, $Z \perp E$). However, this forces predicted label distribution $P(Y|Z)$ to be the same across all the environments~\citep{zhao2019learning}. \citet{long2018conditional,li2018deep,combes2020domain} extends the invariance criterion by conditioning on the label in order to address the label shift issue of DANN.
Invariant risk minimization (IRM)~\citep{arjovsky2019invariant} seeks a different notion of invariance. Instead of aligning distributions of $Z$, IRM requires that the predictor $f$ operating on $Z=\phi(X)$ is simultaneously optimal across different environments. The associated independence is $Y \perp E \;\vert\; Z$. Various work~\citep{krueger2020out,chang2020invariant} has sought to extend IRM. We focus on the structured setting, where most of the environments can uniquely specify $X$ in the training set. As a result, $E$ would act similarly to $X$. In the extreme case, the IRM principle reduces to $Y \perp X \;\vert\; Z$, which is not the desired invariance criterion. We propose to address this issue by introducing domain perturbation to adaptively highlight the structured variation.

\textbf{Domain generalization} These methods seek to learn models that generalize to new domains~\citep{muandet2013domain,ghifary2015domain,motiian2017unified,li2017deeper,li2018domain}. Domain generalization methods can be roughly divided into three categories: \emph{domain adversarial training}~\citep{ganin2016domain,tzeng2017adversarial,long2018conditional}, 
\emph{meta-learning}~\citep{li2018learning,balaji2018metareg,li2019episodic,li2019feature,dou2019domain} and \emph{domain augmentation}~\citep{shankar2018generalizing,volpi2018generalizing}. Our method resembles meta-learning based methods in that we create held-out environments to simulate domain shift during training. However, our objective seeks to reduce the regret between predictors trained with or without access to the held-out environments.

Existing domain generalization benchmarks assume that each domain contains sufficient amounts of data. We focus on a different setting where most of the environments contain only few (or single) examples since they are defined by structured descriptors. This setting often arises in chemical and biological applications (see Figure~\ref{fig:motivation}). Similar to data augmentation method in \citet{shankar2018generalizing}, our structured RGM also creates perturbed examples based on domain-guided perturbations. However, our method operates over learned representations since our inputs are discrete. Moreover, the perturbed examples are only used to regularize the feature extractor $\phi$ via the regret term.

\section{Regret minimization}
\label{sec:method}

To introduce our method, we start with a standard setting where the training set $\mathcal{D}$ is comprised of $n$ environments $\gE = \set{E_1,\cdots,E_n}$~\citep{arjovsky2019invariant}. Each environment $E_i$ consists of examples $(x,y)$ randomly drawn from some distribution $\gP_i$. 
Assuming that new environments we may encounter at test time exhibit similar variability as the training environments, our goal is to train a model that generalizes to such new environments $E_{\mathrm{test}}$. Suppose our model consists of two components $f\circ \phi$, where the predictor $f$ operates on the feature extractor $\phi$. Let $\loss^e(f \circ \phi) = \sum_{(x,y)\in E_e} \ell(y, f(\phi(x)))$ be its empirical loss in environment $E_e$ and $\loss(f \circ \phi) = \sum_e \loss^e(f \circ \phi)$. IRM learns $\phi$ and $f$ such that $f$ is simultaneously optimal in all training environments:
\begin{equation}
    \min_{\phi,f} \; \mathcal{L}(f \circ \phi) \qquad \mathrm{s.t.}\;\; \forall e: f \in \argmin_h \mathcal{L}^e(h \circ \phi)
\end{equation} 
One possible way to solve this objective is through Lagrangian relaxation: 
\begin{equation}
    \min_{\phi,f} \mathcal{L}(f \circ \phi) + \sum_e \lambda_e \big(\mathcal{L}^e(f \circ \phi) - \min_h \mathcal{L}^e(h \circ \phi)\big)
\end{equation}
The regularizer $\mathcal{L}^e(f \circ \phi) - \min_h \mathcal{L}^e(h \circ \phi)$ measures the performance gap between $f$ and the best predictor $\hat{h}\in F_e(\phi)  = \argmin_h \mathcal{L}^e(h \circ \phi)$ specific to environment $E_e$. Note that both $f$ and $\hat{h}$ are trained and evaluated on examples from environment $E_e$. This motivates us to replace the regularizer with a predictive regret. 
Specifically, for each environment $E_e$, we define the associated regret $\mathcal{R}^e(\phi)$ as the difference between the losses of two auxiliary predictors trained \emph{with} and \emph{without} access to examples $(x,y) \in E_e$:
\begin{equation}
    \mathcal{R}^e(\phi) = \loss^e(f_{-e} \circ \phi) - \min_{h\in\gF} \loss^e (h \circ \phi) = \loss^e(f_{-e} \circ \phi) - \loss^e (f_e \circ \phi)\label{eq:regret}
\end{equation}
where the two auxiliary predictors are obtained from (assuming $\gF$ is bounded and closed):
\begin{equation}
    f_e \in  F_e(\phi) = \argmin_{h \in \gF} \mathcal{L}^e(h \circ \phi) \qquad f_{-e} \in  F_{-e}(\phi) = \argmin_{h \in  \gF} \sum_{k \neq e} \mathcal{L}^{k}(h \circ \phi)
    \label{eq:auxpred}
\end{equation}
The \emph{oracle predictor} $f_e$ is trained on environment $E_e$, while $f_{-e}$ uses the rest of the environments $\gE\backslash\set{E_e}$ for training but is tested on $E_e$. Note that $\mathcal{R}^e(\phi)$ does not depend on the predictor $f$ we are seeking to estimate; it is a function of the representation $\phi$ as well as the two auxiliary predictors $f_{-e}$ and $f_{e}$. For notational simplicity, we have omitted $R^e(\phi)$'s dependence on $f_{-e}$ and $f_e$.
Since both predictors are evaluated on the same set of training examples in $E_e$, we immediately have
\begin{prop}
The regret $\gR^e(\phi)$ is always non-negative for any representation $\phi$. 
 \label{thm:regret}
\end{prop}
The proof is straightforward since $f_e$ is the minimizer of $\loss^e (f' \circ \phi)$ and both $f_e$ and $f_{-e}$ are drawn from the same parametric family $\gF$.
The overall regret $\mathcal{R}(\phi) = \sum_e \mathcal{R}^e(\phi)$ expresses our stated goal of finding a representation $\phi$ that generalizes to each held-out environment. Our regret minimization (RGM) objective then balances the empirical loss against the regret, finding $\phi$ and $f$ that minimize:
\begin{equation}
    \mathcal{L}_{\mathrm{RGM}} = \loss(f \circ \phi) + \lambda \sum_e\nolimits \mathcal{R}^e(\phi) \label{eq:rgm}
\end{equation}

\begin{figure}
\centering
\includegraphics[width=\textwidth]{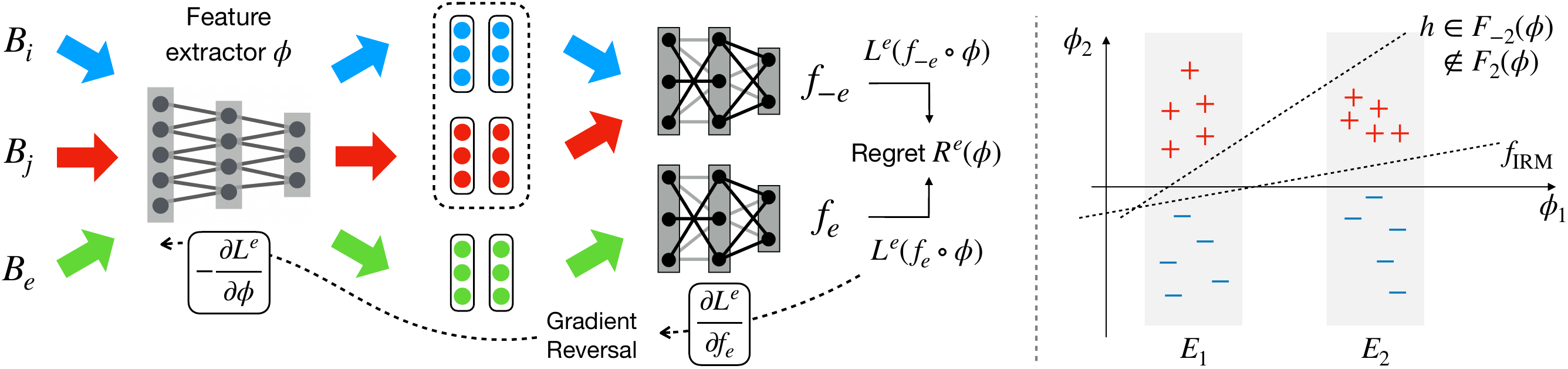}
\vspace{-10pt}
\caption{\emph{Left}: In the backward pass, the gradient of $\loss^e(f_e\circ \phi)$ goes through a gradient reversal layer~\citep{ganin2016domain} which negates the gradient during back-propagation. \emph{Right}: An example where environments are generated by different translations of $X_1$. For the identity mapping $\phi(X)=(X_1,X_2)$, there exists a predictor $f_\mathrm{IRM}$ which is simultaneously optimal in all environments. In contrast, $\phi$ is not feasible under RGM because there is a linear classifier $h \in F_{-2}(\phi)$ but $h \not\in F_2(\phi)$.}
\label{fig:rgm}
\end{figure}

\subsection{Comparison with IRM}
Compared to IRM, the proposed RGM objective imposes a stronger constraint on $\phi$ since $f_{-e}$ is not trained on $E_e$. To show this formally, let $F_e(\phi), F_{-e}(\phi)$ be the set of optimal predictors in $E_e$ and $\gE\backslash\set{E_e}$ respectively as defined in Eq.(\ref{eq:auxpred}).
Since $\mathcal{R}^e(\phi)=0 \Leftrightarrow f_{-e} \in F_e(\phi)$ and $f_{-e}$ is chosen arbitrarily from $F_{-e}(\phi)$, the constrained form of the RGM objective can be stated as
\begin{equation}
    \min_{\phi,f} \; \mathcal{L}(f \circ \phi) \qquad \mathrm{s.t.}\;\; \forall e: F_{-e}(\phi) \subseteq F_e(\phi)
\end{equation}
The analogous IRM constraints are $f \in \cap_e F_e(\phi)$ and $\cap_e F_e(\phi)\neq \emptyset$.
Suppose both IRM and RGM constraints are feasible and let $\loss_{\mathrm{IRM}}^*,\allowbreak \loss_{\mathrm{RGM}}^*$ be their optimal loss respectively.
Consider the set of optimal features under both objectives:
\begin{eqnarray} 
\Phi_{\mathrm{IRM}} &=& \set{\phi \;|\; \min_{f \in \cap_e F_e(\phi)} \mathcal{L}(f \circ \phi) = \loss_{\mathrm{IRM}}^*,\; \cap_e F_e(\phi)\neq \emptyset} \\
\Phi_{\mathrm{RGM}} &=& \set{\phi \;|\; \min_{f\in\gF} \mathcal{L}(f \circ \phi) = \loss_{\mathrm{RGM}}^*,\; \forall e: F_{-e}(\phi) \subseteq F_e(\phi)}
\end{eqnarray}
\begin{prop}
Assuming two environments, if $\loss_{\mathrm{RGM}}^* = \loss_{\mathrm{IRM}}^*$, then $\Phi_{\mathrm{RGM}} \subseteq \Phi_{\mathrm{IRM}}$. The converse  $\Phi_{\mathrm{IRM}} \subseteq \Phi_{\mathrm{RGM}}$ does not hold in general. 
\label{thm:rgm}
\end{prop}
While limited to two environments, the proposition suggests that RGM imposes stronger constraints on $\phi$.
Figure~\ref{fig:rgm} shows a counterexample illustrating that $\Phi_{\mathrm{IRM}} \not\subseteq \Phi_{\mathrm{RGM}}$. Suppose there are two environments generated by translation of $X_1$ and the true hypothesis is $\mathbb{I}[X_2 > 0]$. The identity mapping $\phi(X)=(X_1,X_2)$ is not translation invariant, but $\phi \in \Phi_{\mathrm{IRM}}$ because there exists a predictor $f_\mathrm{IRM}$ that is simultaneously optimal in all environments. On the other hand, $\phi$ is not feasible under RGM because there is a linear classifier $h \in F_{-2}(\phi)$ that is optimal in $E_1$ but suboptimal in $E_2$, violating the RGM constraint $F_{-2}(\phi) \subseteq F_2(\phi)$. Thus $\phi \not\in \Phi_{\mathrm{RGM}}$. 

To see why it would be helpful to add a stronger constraint on $\phi$, consider the following data generation process where the environment $e$ can be inferred from $x$ alone:
\begin{eqnarray}
    p(x,y,e) = p(e)p(x|e)p(y|x,e); \qquad p(y|x,e) = p(y|x,e(x)) \label{eq:A2}
\end{eqnarray}
For molecules and proteins, this assumption is often valid because the environment labels (scaffolds, protein families) typically depend on $x$ only. We call $\phi$ \emph{label-preserving} if it retains all the information about the label: $p(y | \phi(x)) = p(y | x, e)$. Such representation may not generalize to new environments given the dependence on $e$ through $\phi$. However, we can show that for any label-preserving $\phi$, its associated ERM optimal predictor also satisfies the IRM constraints:
\begin{prop}
For any label-preserving $\phi$ with $p(y | \phi(x)) = p(y | x, e)$, its associated ERM optimal predictor $f^*$ satisfies the IRM constraint. Moreover, if $\phi \in \Phi_{\mathrm{IRM}}$, $f^* \circ \phi$ is optimal under IRM.
\label{thm:irm}
\end{prop}
While IRM constraints are vacuous for any label-preserving $\phi$, this is not necessarily the case with RGM constraints. Consider, for example, the counterexample in Figure~\ref{fig:rgm}. The identity mapping $\phi(X)=(X_1,X_2)$ is label-preserving since it retains all the input information. However, $\phi$ is infeasible under RGM.

\subsection{Structured Environments}
\label{sec:srgm}
\begin{figure}[t]
    \centering
    \includegraphics[width=\textwidth]{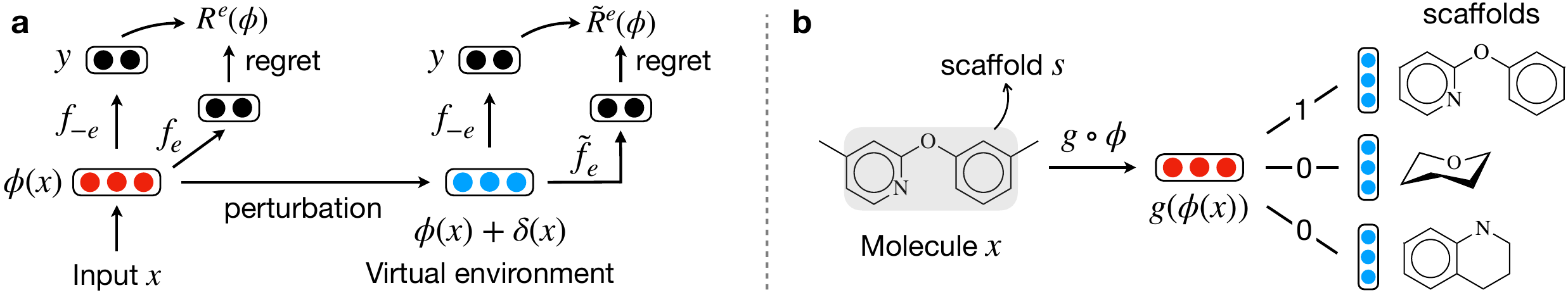}
    \vspace{-14pt}
    \caption{a) Structured RGM: we introduce additional oracle predictors $\tilde{f}_{e}$ for the perturbed inputs; b) In molecule tasks, the scaffold classifier $g$ is trained by negative sampling.}
    \label{fig:srgm}
\end{figure}

Now let us consider a more challenging setting, where we have another set of environments defined via structured descriptors besides the standard environments $\set{E_1,\cdots,E_n}$. Formally, the structured environments comes in the form $\gD=\{(x_i,y_i,s_i)\}$, where $s_i$ is a structured environment descriptor of $(x,y)$. For instance, in molecule property prediction, $s_i$ is defined as a Murcko scaffold (i.e., subgraph) of molecule $x_i$. 
It is hard to turn the scaffolds into standard environments because scaffolds are structured descriptors and they often uniquely identify each molecule in the training set (Figure~\ref{fig:motivation}). Creating single-example environments is not helpful as the model would see any change from one example to another as scaffold variation.

Alternatively, we can describe scaffold variation by perturbation in the representation $\phi$. The idea is to create a perturbed instance $\tilde{x}_i$ for each example $(x_i,y_i,s_i)$ so that the difference between $x_i$ and $\tilde{x}_i$ highlights how scaffold information has changed in the representation. 
Specifically, the perturbation $\delta(x_i)$ is defined through a parametric scaffold classifier $g$ built on top of the representation $\phi$.\footnote{Our method is introduced using scaffolds as examples. It can be applied to other structured environments like protein families by simply replacing the scaffold classifier with a protein family classifier.}
The associated scaffold classification loss is $\ell(s_i, g(\phi(x_i)))$. 
Given that our inputs are discrete, we define the perturbation $\delta$ as the gradient with respect to the continuous representation $\phi$:
\begin{equation}
    \phi(\tilde{x_i}) \coloneqq \phi(x_i) + \delta(x_i) = \phi(x_i) + \alpha \nabla_{z} \ell(s_i, g(z)) |_{z=\phi(x_i)} \label{eq:perturb}
\end{equation}
where $\alpha$ is a step size parameter. The perturbation is specifically designed to contain less information about the scaffold $s_i$, and we require that the model should not be affected by this variation in the representation. Since these perturbations introduce additional simulated test scenarios that we wish to generalize to, we propose to regularize our model also based on regret associated with perturbed inputs.
Similar to Eq.(\ref{eq:regret}), the regret corresponding to perturbed inputs is defined as
\begin{eqnarray}
    \mathcal{R}^e(\phi + \delta) &=& \loss^e(f_{-e} \circ (\phi + \delta)) - \min_h\nolimits \loss^e(h \circ (\phi + \delta)) \\
    \loss^e(h \circ (\phi + \delta)) &=& \sum_{(x_i,y_i) \in E_e}\nolimits \ell\big(y_i, h(\phi(x_i) + \delta(x_i))\big)
\end{eqnarray}
which introduces a new oracle predictor $\tilde{f}_{e} = \argmin_h \loss^e (h \circ (\phi + \delta))$ for each environment $E_e$ (see Figure~\ref{fig:srgm}a). 
Note that $f_{-e}$ is the same auxiliary predictor as before. It minimizes a separate objective $\mathcal{L}^{-e}(f_{-e} \circ \phi)$, which does \emph{not} include the perturbed examples.

The structured RGM objective $\mathcal{L}_{\mathrm{SRGM}}$ augments the basic RGM with additional regret terms as well as the scaffold classification loss $\mathcal{L}_g(g \circ \phi)$:
\begin{eqnarray}
    \mathcal{L}_{\mathrm{SRGM}} &=& \loss(f \circ \phi) + \lambda_g \mathcal{L}_g(g \circ \phi) + \lambda \sum_e\nolimits \sum_{\psi \in \set{0, \delta}}\nolimits \mathcal{R}^e(\phi + \psi) \\
    \mathcal{L}_g(g \circ \phi) &=& \sum_{(x,y,s) \in  \mathcal{D}}\nolimits \ell\big(s, g(\phi(x))\big)
\end{eqnarray}
The forward pass of SRGM is shown in Algorithm~\ref{alg:srgm}.
Since $s$ is a structured object with a large number of possible values, we train the classifier $g$ with negative sampling (Figure~\ref{fig:srgm}b).
Note that $\phi$ is also updated to partially optimize $\mathcal{L}_g$. This is necessary to ensure that the scaffold classifier operating on $\phi$ has enough information to introduce a reasonable gradient perturbation $\delta(x)$. This trade-off keeps some scaffold information in $\phi$ while ensuring, via the associated regret terms, that this information is not strongly relied upon. The effect of this design choice is studied in the appendix.

\begin{algorithm}[t]
\begin{algorithmic}[1]
\caption{Structured RGM: Forward Pass}
\label{alg:srgm}
\STATE Sample two minibatches $B_e$ from environment $E_e$
\FOR{each environment $E_e\in \gE$}
\STATE Compute scaffold classification loss $\mathcal{L}_g(g \circ \phi)$ over $B_e$.
\STATE Construct perturbed examples $\tilde{B}_e$ via gradient perturbation (see Eq.(\ref{eq:perturb})) .
\STATE Compute empirical loss $\mathcal{L}(f \circ \phi)$ on $B_e$.
\STATE Compute auxiliary predictor loss $\mathcal{L}^{-e}(f_{-e} \circ \phi)$ on $B_{-e}$.
\STATE Compute oracle predictor losses $\mathcal{L}^e(f_e \circ \phi)$ and $\mathcal{L}(\tilde{f}_{e} \circ \phi, \tilde{E}_e)$ on $B_e$ and $\tilde{B}_e$.
\STATE Compute regret terms $\mathcal{R}^e(\phi),\mathcal{R}^e(\phi + \delta)$ on $B_e$ and $\tilde{B}_e$.
\ENDFOR
\end{algorithmic}
\end{algorithm}

\subsection{Optimization } 
The standard RGM objective in Eq.(\ref{eq:rgm}) can be viewed as finding a stationary point of a multi-player game between $f$, $\phi$ as well as the auxiliary predictors $\{f_{-e}\}$ and $\{f_e\}$. Our predictor $f$ and representation $\phi$ find their best response strategies by minimizing
\begin{equation}
    \min_{f,\phi}\big\{\loss(f \circ \phi) + \lambda \sum_e\nolimits \big(\loss^e(f_{-e} \circ \phi) - \loss^e (f_e \circ \phi)\big)\big\} \label{eq:rgm1}
\end{equation}
while the auxiliary predictors minimize 
\begin{equation}
\min_{f_{-e}} \mathcal{L}^{-e}(f_{-e} \circ \phi)
\;\;\;\text{ and }\;\;\;
\min_{f_e} \mathcal{L}^{e}(f_{e} \circ \phi)\;\;\;
\forall e
\label{eq:rgm2}
\end{equation}
This multi-player game can be optimized by stochastic gradient descent. Since $f_e$ and $\phi$ optimizes $\loss^e(f_e \circ \phi)$ in opposite directions, we introduce a gradient reversal layer~\citep{ganin2016domain} between $\phi$ and $f_e$. This allows us to update all the players in a single forward-backward pass (see Figure~\ref{fig:rgm}). In each step, we simultaneously update all the players by 
\begin{align*}
    f &\leftarrow f - \eta \nabla_f \loss(f \circ \phi) &&
    \;\phi \leftarrow \phi - \eta \nabla_\phi \loss(f \circ \phi) - \eta \lambda  \sum_e\nolimits \nabla_\phi \mathcal{R}^e(\phi) \\
    f_{-e} &\leftarrow f_{-e} - \eta \nabla \mathcal{L}^{-e}(f_{-e} \circ \phi) &&
    f_e \leftarrow f_e - \eta \nabla \mathcal{L}^{e}(f_{e} \circ \phi) \quad \forall e
\end{align*}
where $\mathcal{L}^{-e}(f_{-e} \circ \phi) = \sum_{k\neq e} \mathcal{L}^{k}(f_{-e} \circ \phi)$. In each step, we sample minibatches $B_1,\cdots,B_n$ from each environment $E_1,\cdots,E_n$. The loss $\loss(f \circ \phi)$ is computed over all the minibatches $\bigcup_k B_k$, while $\loss^{-e}(f_{-e} \circ \phi)$ is computed over minibatches $B_{-e} = \bigcup_{k\neq e} B_k$. The regret term $R^e(\phi)$ is evaluated based on examples in $B_e$ only.

For structured RGM, its optimization rule is analogous to RGM, with additional gradient updates for the oracle predictors $\tilde{f}_e$ and scaffold classifier $g$ (see Appendix~\ref{sec:srgmopt}). While the perturbation $\delta$ is defined on the basis of $\phi$ and $g$, we do not include the dependence during back-propagation as incorporating this higher order gradient does not improve our empirical results.
\section{Experiments}
\label{sec:experiment}
Our methods (RGM and SRGM) are evaluated on real-world applications such as molecular property prediction, protein homology and stability prediction. Our baselines include:
\begin{itemize}[leftmargin=*,topsep=0pt,itemsep=0pt]
\item Standard empirical risk minimization (ERM) trained on aggregated environments;
\item Domain adversarial training methods including DANN~\citep{ganin2016domain} and CDAN~\citep{long2018conditional}, which seek to learn domain-invariant features;
\item IRM~\citep{arjovsky2019invariant} requiring the model to be simultaneously optimal in all environments;
\item MLDG~\citep{li2018learning}, a meta-learning method which simulates domain shift by dividing training environments into meta-training and meta-testing;
\item CrossGrad~\citep{shankar2018generalizing} which augments the training set with domain-guided perturbations of inputs. Since our inputs are discrete, we perform perturbation on the representation instead.
\end{itemize}

\begin{table}[t]
\centering
\caption{Mean absolute error (MAE) on the QM9 dataset. Models are trained on molecules with no more than 7 atoms and tested on molecules with 9 atoms. Due to space limit, we only show standard deviation for the top three methods in subscripts. We highlight the best method in each setup.}
\small
\vspace{-3pt}
\begin{tabular}{lcccccc|cc}
\hline
& \multicolumn{6}{c|}{Categorical environments} & \multicolumn{2}{c}{Scaffold environments}\Tstrut\Bstrut \\
\hline
Property & ERM & DANN & CDAN & IRM & MLDG & RGM & CrossGrad & SRGM \Tstrut\Bstrut \\
\hline
mu & 0.658 & 0.655 & 0.655 & 0.690 & \textbf{0.654} & 0.656\textsubscript{(.004)} & 0.664\textsubscript{(.001)} & 0.666\textsubscript{(.005)} \Tstrut\Bstrut \\
alpha & 13.08 & 13.17 & 13.19 & 13.16 & 14.13 & \textbf{12.99}\textsubscript{(.028)} & 12.79\textsubscript{(.379)} & \textbf{11.54}\textsubscript{(.777)} \Tstrut\Bstrut \\
HOMO & 0.008 & 0.008 & 0.008 & 0.009 & 0.008 & 0.008\textsubscript{(.000)} & 0.008\textsubscript{(.000)} & 0.009\textsubscript{(.000)} \Tstrut\Bstrut \\
LUMO & 0.011 & 0.011 & 0.011 & 0.011 & 0.011 & \textbf{0.010}\textsubscript{(.000)} & 0.011\textsubscript{(.000)} & 0.013\textsubscript{(.000)} \Tstrut\Bstrut \\
gap & 0.014 & 0.013 & 0.014 & 0.015 & 0.014 & \textbf{0.012}\textsubscript{(.001)} & 0.014\textsubscript{(.001)} & 0.016\textsubscript{(.001)} \Tstrut\Bstrut \\
R2 & 352.8 & 355.7 & 357.3 & 368.6 & 381.2 & \textbf{328.4}\textsubscript{(11.2)} & 351.7\textsubscript{(11.0)} & \textbf{279.9}\textsubscript{(29.6)} \Tstrut\Bstrut \\
ZPVE & 0.025 & 0.024 & 0.025 & 0.025 & 0.026 & \textbf{0.022}\textsubscript{(.000)} & 0.024\textsubscript{(.001)} & \textbf{0.019}\textsubscript{(.001)} \Tstrut\Bstrut \\
Cv & 5.336 & 5.351 & 5.369 & 5.327 & 5.756 & \textbf{4.860}\textsubscript{(.228)} & 5.235\textsubscript{(.176)} & \textbf{3.909}\textsubscript{(.420)} \Tstrut\Bstrut \\
U0 & 67.18 & 67.57 & 67.34 & 67.67 & 71.83 & \textbf{60.25}\textsubscript{(2.62)} & 63.82\textsubscript{(1.82)} & \textbf{51.32}\textsubscript{(4.51)} \Tstrut\Bstrut \\
U & 66.67 & 67.00 & 67.24 & 68.55 & 71.60 & \textbf{58.74}\textsubscript{(2.51)} & 64.30\textsubscript{(1.47)} & \textbf{51.54}\textsubscript{(5.09)} \Tstrut\Bstrut \\
H & 67.00 & 67.39 & 67.27 & 68.23 & 71.47 & \textbf{59.72}\textsubscript{(2.23)} & 64.39\textsubscript{(2.19)} & \textbf{50.17}\textsubscript{(2.56)} \Tstrut\Bstrut \\
G & 65.92 & 65.95 & 66.02 & 68.16 & 70.70 & \textbf{59.40}\textsubscript{(2.12)} & 64.63\textsubscript{(1.12)} & \textbf{51.23}\textsubscript{(6.13)} \Tstrut\Bstrut \\
\hline
\end{tabular}
\label{tab:qm9}
\vspace{-5pt}
\end{table}

\subsection{Molecular property Prediction}

\textbf{Data } The training data consists of $\{(x_i,y_i,s_i)\}$, where $x_i$ is a molecular graph, $y_i$ is its property and $s_i$ is its scaffold. We adopt four datasets from the MoleculeNet benchmark~\citep{wu2018moleculenet}:
\begin{itemize}[leftmargin=*,topsep=0pt,itemsep=0pt]
\item QM9 is a regression dataset of 134K compounds with 12 properties related to quantum chemistry. We split the dataset based on number of atoms: our training set contains molecules with no more than 7 atoms; our validation and test set consist of molecules with 8 and 9 atoms respectively. This setup is much harder than random split as it requires models to extrapolate to new chemical space.
\item HIV, Tox21 and BBBP are three classification datasets related to medicinal chemistry. The three datasets contains 36K, 7K and 2K molecules respectively. The training and test sets are built by scaffold splitting. To measure extrapolation, we sort the scaffolds by their molecular weight and put the top 10\% largest scaffolds in the test set. The training set contains 80\% of the smallest scaffolds and the rest 10\% scaffolds form the validation set (details in the appendix).
\end{itemize}

\textbf{Setup } The training set consists of multiple environments based on the Murcko scaffold~\citep{bemis1996properties}. Since most of our baselines cannot utilize structural information of the environments (i.e., scaffolds), we consider two evaluation setup for fair comparison:
\begin{itemize}[leftmargin=*,topsep=0pt,itemsep=0pt]
\item \emph{Categorical environments}: We cluster all the training environments into two environments $E_0,E_1$. For QM9, $E_1$ contains molecules with 7 atoms. For other datasets, $E_1$ contains the top 25\% largest scaffolds. Under this setup, DANN, CDAN, IRM and MLDG are comparable with RGM. 
\item \emph{Scaffold environments}: We compare SRGM with CrossGrad in this setup. Both methods utilizes scaffold information via gradient perturbation from a scaffold classifier $g$. 
\end{itemize}
Following~\citet{wu2018moleculenet}, we report mean absolute error (MAE) for QM9 and AUROC for the HIV, Tox21 and BBBP. All the results are averaged across five independent runs.

\textbf{Model } The molecule encoder $\phi$ is a graph convolutional network~\citep{yang2019analyzing} which translates a molecular graph into a continuous vector. The predictor $f$ is a two-layer MLP that takes $\phi(x)$ as input and predicts the label. The scaffold classifier $g$ is also a two-layer MLP trained by negative sampling since scaffold is a combinatorial object with a large number of possible values.
Specifically, for a given molecule $x$ with scaffold $u(x)$, we randomly sample $K$ other molecules and take their associated scaffolds $\{u_k\}$ as negative classes. Details of model architecture and hyper-parameters are discussed in the appendix.

\textbf{Results } Our results on the QM9 dataset are shown in Table~\ref{tab:qm9}. In the categorical setup, RGM outperforms all the baselines (except for property mu), with significant improvement on six properties (R2, Cv, U0, U, H, G) with 7-10\% relative error reduction. 
In the scaffold setup, SRGM outperforms all the baselines on eight properties (out of 12). While CrossGrad utilizes scaffold information, its performance is worse than RGM in general. Compared to RGM, SRGM shows significant error reduction (10-20\%) on seven properties (alpha, R2, Cv, U0, U, H, G). This validates the advantage of exploiting structures of the environments (scaffolds).

\begin{figure}[t]
\centering
\begin{minipage}{0.27\textwidth}
\includegraphics[width=\textwidth]{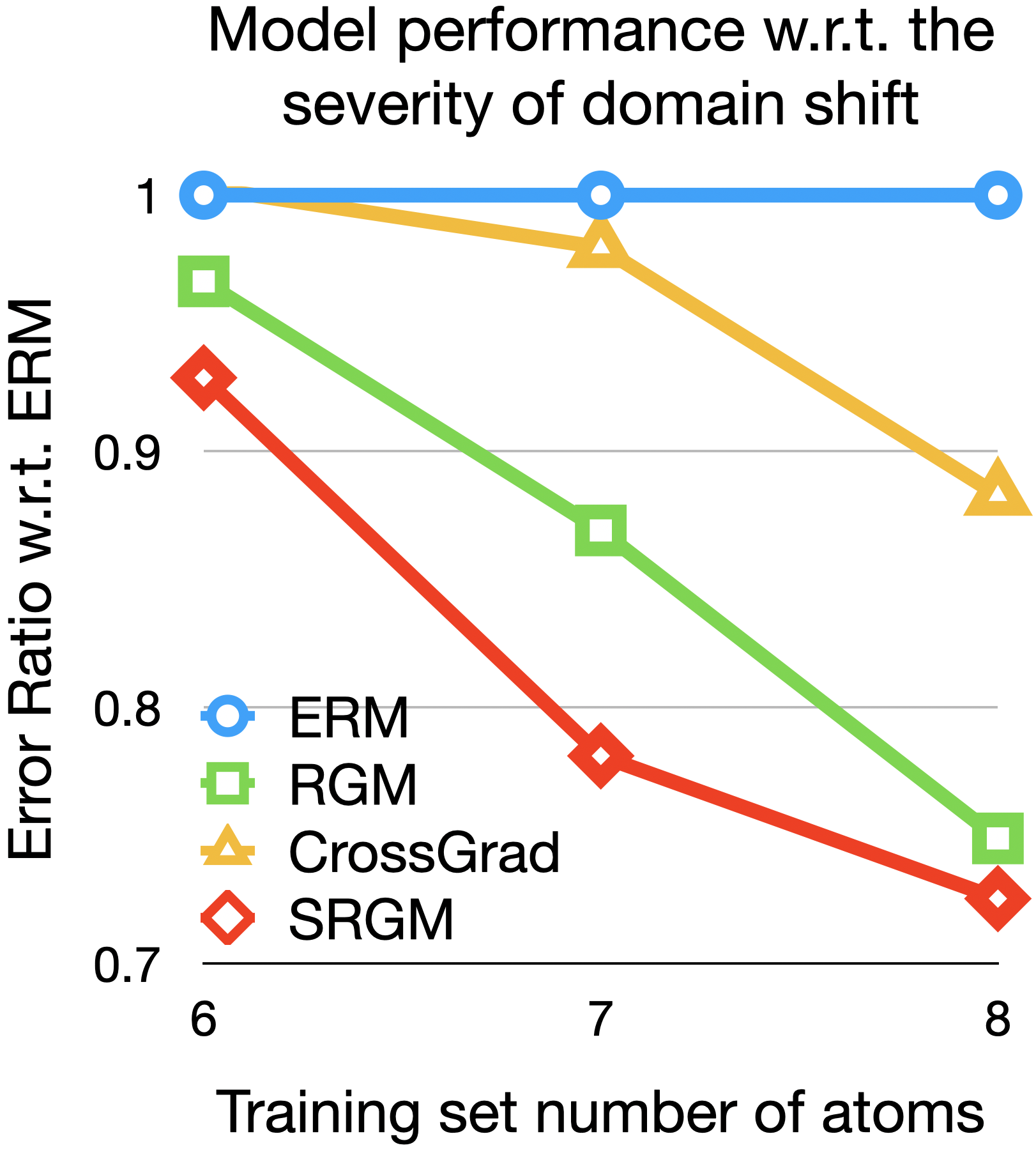}
\end{minipage}
~
\begin{minipage}{0.7\textwidth}
\small
\begin{tabular}{lccc|cc}
\hline
 & HIV & Tox21 & BBBP & HOMO & stability \Tstrut\Bstrut \\
\hline
ERM & 0.614\textsubscript{(.031)} & 0.690\textsubscript{(.008)} & 0.895\textsubscript{(.008)} & 20.9\% & 0.736 \Tstrut\Bstrut \\
DANN & 0.624\textsubscript{(.023)} & 0.680\textsubscript{(.005)} & 0.896\textsubscript{(.002)} & 22.3\% & 0.770  \Tstrut\Bstrut \\
CDAN & 0.613\textsubscript{(.055)} & 0.682\textsubscript{(.004)} & 0.896\textsubscript{(.005)} & 21.9\% & 0.750  \Tstrut\Bstrut \\
IRM & 0.637\textsubscript{(.043)} & 0.686\textsubscript{(.003)} & 0.896\textsubscript{(.009)} & 21.0\% & 0.723 \Tstrut\Bstrut \\
MLDG & 0.639\textsubscript{(.057)} & 0.686\textsubscript{(.009)} & 0.896\textsubscript{(.003)} & 22.0\% & 0.754 \Tstrut\Bstrut \\
RGM & 0.644\textsubscript{(.028)} & 0.685\textsubscript{(.004)} & 0.895\textsubscript{(.007)} & \textbf{23.4\%} & \textbf{0.787} \Tstrut\Bstrut \\
\hline
CrossGrad & 0.708\textsubscript{(.035)} & 0.694\textsubscript{(.006)} & 0.902\textsubscript{(.002)} & 20.9\% & 0.662 \Tstrut\Bstrut \\
SRGM & \textbf{0.735}\textsubscript{(.015)} & \textbf{0.701}\textsubscript{(.003)} & \textbf{0.911}\textsubscript{(.003)} & \textbf{23.8\%} & \textbf{0.793} \Tstrut\Bstrut \\
\hline
\end{tabular}
\captionlistentry[table]{A table beside a figure} \label{tab:rest}
\end{minipage}
\captionsetup{labelformat=andtable}
\caption{\emph{Left}: Ablation study on QM9 with various levels of domain shift. Models are trained on three different sets: molecules with less than 8, 7 and 6 atoms (increasing domain shift). \emph{Right}: Results on molecule and protein datasets. CrossGrad and SRGM operate on the structured environments (scaffolds or topology), while RGM and others operate on categorical environments.}
\vspace{-5pt}
\label{fig:ablation}
\end{figure}

Results on the classification datasets are shown in Table~\ref{tab:rest}. In the categorical setup, RGM performs similarly to other baselines (within one standard deviation). In the scaffold setup, SRGM shows clear improvement on all three datasets, with a significant accuracy increase in the HIV dataset (0.735 v.s. 0.644). This further confirms the importance of exploiting the structure of domain shift.

\textbf{Ablation study } We conduct additional experiments to study the performance of RGM/SRGM with respect to the severity of domain shift. Fixing the test set to molecules with 9 atoms, we construct three progressively harder training sets: molecules with no more than 8, 7 and 6 atoms. We report the MAE ratio (averaged over 12 properties) between SRGM/RGM/CrossGrad and ERM. As shown in Figure~\ref{fig:ablation}, SRGM consistently outperforms CrossGrad and RGM across different setups.

\subsection{Protein Modeling}

\textbf{Data } The dataset consists of pairs $\{(x_i,y_i,s_i)\}$, where $x_i$ is a protein represented as sequence of amino acid characters and $y_i$ denotes its property. We consider two datasets used in \citet{tape2019}:
\begin{itemize}[leftmargin=*,topsep=0pt,itemsep=0pt]
\item \emph{Homology prediction} (HOMO)~\citep{scop}: The dataset consists of 12K for training, 736 for validation and 718 for testing, which are split by protein superfamilies (evolutionary groups). This requires models to generalize across large evolutionary gaps. There are 1823 protein superfamilies in total, with around 1200 of them having less than 10 instances in the training set.
\item \emph{Stability prediction}~\citep{rocklin2017}: The dataset has 54K for training, 2.4K for validation and 13K for testing. The train set contains proteins sampled broadly across sequence space, while the test set contains Hamming distance-1 neighbors of most stable proteins. The dataset contains 1218 topology groups, with around 1200 of them having no more than two instances.
\end{itemize}

\textbf{Setup } Each environment $s_i$ corresponds to a protein superfamily or topology. Since most environments contain very few examples, we cluster them into two environments $E_0,E_1$. For the homology task, $E_1$ contains all superfamilies with less than 10 proteins and $E_0 = E - E_1$. For the stability task, we sort the topology groups by their frequency and divide them evenly into $E_0$ and $E_1$. SRGM and CrossGrad use gradient perturbation from a protein superfamily/topology classifier $g$, while RGM and other baselines are trained on the clustered environments $E_0, E_1$.

\textbf{Model } Our protein encoder $\phi$ is a pre-trained BERT model~\citep{tape2019}. The predictor $f$ is a linear function that takes $\phi(x)$ as input and predicts its fold label or stability score. The superfamily and topology classifier $g$ is a two-layer MLP. The hyperparameters are listed in the appendix.

\textbf{Results } Following \citet{tape2019}, we report the top-1 accuracy for homology prediction and Spearman ranking correlation $\rho$ for stability prediction. Our ERM baseline matches their Transformer performance. Both RGM and SRGM outperforms all the baselines in both tasks (homology: 23.8\% v.s. 22.3\%; stability: 0.793 v.s. 0.770). The difference between RGM and SRGM is relatively smaller compared to the molecule domain.

\section{Conclusion}
In this paper, we propose regret minimization for generalization across structured biomedical domains such as molecular scaffolds or protein families. We seek to find a representation that enables the predictor to compete against an oracle with hindsight access to unseen domains. Our method significantly outperforms all baselines on real-world biomedical tasks.

\bibliography{iclr2021_conference}
\bibliographystyle{iclr2021_conference}

\appendix
\newpage
\appendix

\section{Technical Details}
\subsection{Proof of Proposition~\ref{thm:regret}}
Note that $\mathcal{L}^e(f \circ \phi)$ is defined on a set of fixed examples in $E_e$. Since $f_e \in \argmin_{f' \in \gF}\mathcal{L}^e(f' \circ \phi)$ and $f_e,f_{-e}$ are in the same parametric family $\gF$, we have $\mathcal{R}^e(\phi) = \loss^e(f_{-e} \circ \phi) - \loss^e (f_e \circ \phi) \geq 0$.
\subsection{Proof of Proposition~\ref{thm:rgm}}
\begin{proof}
Consider any representation $\phi^*\in \Phi_{\mathrm{RGM}}$. When there are only two environments $\set{E_1, E_2}$, we have $F_{-2}(\phi^*) = F_1(\phi^*)$ and $F_{-1}(\phi^*) = F_2(\phi^*)$ by definition. Thus the RGM constraint implies
$$
F_2(\phi^*) = F_{-1}(\phi^*)\subseteq F_1(\phi^*) \qquad F_1(\phi^*) = F_{-2}(\phi^*)\subseteq F_2(\phi^*)
$$
Therefore $F_1(\phi^*) = F_2(\phi^*)$. Since the loss function is non-negative and $\gF$ is bounded and closed, $F_1(\phi^*) \neq \emptyset$. Thus, $\cap_e F_e(\phi^*) = F_1(\phi^*) \neq \emptyset$.
Now consider any $f \in \cap_e F_e(\phi^*)$. By definition,
$$
\forall e: \loss^e(f \circ \phi^*) \leq \min_{h\in\gF} \loss^e(h \circ \phi^*)
$$
By summing the above inequality over all environments, we have
$$
\sum_e \loss^e(f \circ \phi^*) \leq \sum_e \min_{h\in\gF} \loss^e(h \circ \phi^*) \leq \min_{h\in\gF} \sum_e \loss^e(h \circ \phi^*)
$$
Since $\sum_e \loss^e(f \circ \phi^*) =  \loss(f \circ \phi^*)$, the above inequality implies
$$
\loss(f \circ \phi^*) \leq \min_{h\in\gF} \loss(h \circ \phi^*) = \loss^*_{\mathrm{RGM}} = \loss^*_{\mathrm{IRM}}
$$
Thus, $f \circ \phi^*$ is an optimal solution under IRM and $\phi^* \in \Phi_{\mathrm{IRM}}$.
\end{proof}

\subsection{Proof of Proposition~\ref{thm:irm}}

\begin{proof}
Let us recall our assumption of the data generation process: 
$$
    p(x,y,e) = p(e)p(x|e)p(y|x,e); \qquad p(y|x,e) = p(y|x,e(x))
$$
Under this assumption, we can rephrase the IRM objective as
\begin{align}
    \min_{f,\phi}\quad & \mathbb{E}_e \mathbb{E}_{x|e} \mathbb{E}_{y|x,e} \ell(y, f(\phi(x))) \\
    \mathrm{s.t.} \quad & \mathbb{E}_{x|e} \mathbb{E}_{y|x,e} \ell(y, f(\phi(x))) \leq \min_{f_e} \mathbb{E}_{x|e} \mathbb{E}_{y|x,e} \ell(y, f_e(\phi(x)))  \quad \forall e
\end{align}
Given any label-preserving representation $\phi(x)$, its ERM optimal predictor is 
\begin{equation}
    f^*(\phi(x))=\arg\min_f \mathbb{E}_{y | \phi(x)} \ell(y, f(\phi(x)))
\end{equation}
To see that $f^*$ is ERM optimal, consider
\begin{eqnarray}
\min_{f} \mathbb{E}_e \mathbb{E}_{x|e} \mathbb{E}_{y|x,e} \ell(y, f(\phi(x)))  &\geq& \mathbb{E}_e \mathbb{E}_{x|e} \min_f \mathbb{E}_{y|x,e} \ell(y, f(\phi(x)))  \\
&=& \mathbb{E}_e \mathbb{E}_{x|e} \min_f \mathbb{E}_{y|\phi(x)} \ell(y, f(\phi(x)))  \label{eq:fea1} \\
&=& \mathbb{E}_e \mathbb{E}_{x|e} \mathbb{E}_{y|\phi(x)} \ell(y, f^*(\phi(x))) 
\end{eqnarray}
where Eq.(\ref{eq:fea1}) holds because $\phi(x)$ is label-preserving.
Note that $f^*$ satisfies the IRM constraint because it is simultaneously optimal across all environments:
\begin{eqnarray}
\forall e: \min_{f_e} \mathbb{E}_{x|e}\mathbb{E}_{y|x,e} \ell(y, f_e(\phi(x))) &\geq& \mathbb{E}_{x|e} \min_{f_e} \mathbb{E}_{y|x,e} \ell(y, f_e(\phi(x))) \\
&=& \mathbb{E}_{x|e} \min_{f} \mathbb{E}_{y|\phi(x)} \ell(y, f(\phi(x))) \\
&=& \mathbb{E}_{x|e} \mathbb{E}_{y|\phi(x)} \ell(y, f^*(\phi(x))) \quad 
\end{eqnarray}
Moreover, if $\phi \in \Phi_{\mathrm{IRM}}$ is an optimal representation, $f^* \circ \phi$ is an optimal solution of IRM.
\end{proof}

\subsection{Structured RGM Update Rule}
\label{sec:srgmopt}
Since $\tilde{f}_e$ and $\phi$ optimizes $\mathcal{L}(\tilde{f}_{e} \circ \phi, \tilde{E}_e)$ in different directions, we also introduce a gradient reversal layer between $\phi$ and $\tilde{f}_e$. The SRGM update rule is the following:
\begin{align*}
    \phi &\leftarrow \phi - \eta \nabla_\phi \loss(f \circ \phi) - \eta \lambda_g \nabla_\phi \mathcal{L}_g(g \circ \phi) - \eta \lambda  \sum_e\nolimits\sum_{\psi \in \set{0,\delta}}  \nabla_\phi
    \mathcal{R}^e(\phi + \psi) \\
    f &\leftarrow f - \eta \nabla_f \loss(f \circ \phi) \qquad\quad g \leftarrow g - \eta \nabla_g \mathcal{L}_g(g \circ \phi) \\
    f_e &\leftarrow f_e - \eta \nabla \mathcal{L}^{e}(f_{e} \circ \phi) \qquad
    \tilde{f}_e \leftarrow \tilde{f}_e - \eta \nabla \mathcal{L}(\tilde{f}_{e} \circ (\phi + \delta)) \quad \forall e \\
    f_{-e} &\leftarrow f_{-e} - \eta \nabla \mathcal{L}^{-e}(f_{-e} \circ \phi) \quad \forall e
\end{align*}

\begin{table}[t]
    \centering
    \caption{Dataset statistics} \vspace{-5pt}
    \begin{tabular}{lcccccc}
    \hline
    & QM9 & HIV & Tox21 & BBBP & Homology & Stability \Tstrut\Bstrut \\
    \hline
    Training & 4K & 25243 & 6427 & 1580 & 12.3K & 54K \Tstrut\Bstrut \\
    Validation & 18K & 6352 & 568 & 206 & 736 & 2.4K \Tstrut\Bstrut \\
    Testing & 113K & 3959 & 839 & 256 & 718 & 13K \Tstrut\Bstrut \\
    \hline
    \end{tabular}
    \label{tab:data}
\end{table}

\section{Experimental Details}
\subsection{Molecular property prediction}
\label{sec:appendix_mol}
\textbf{Data } The four property prediction datasets are provided in the supplementary material, along with the training/validation/test splits. The size of each training environment, validation and test set are listed in Table~\ref{tab:data}. The QM9, Tox21 and BBBP dataset are downloaded from \citet{wu2018moleculenet}. The HIV dataset is downloaded from the original source with EC50 measurements.\footnote{\url{https://wiki.nci.nih.gov/download/attachments/158204006/aids_ec50_may04.txt?version=1&modificationDate=1378736563000&api=v2}}
The positive class is defined as molecules with EC50 less than 1$\mu$M.

For the QM9 ablation study, we consider three training sets $\gD_8,\gD_7,\gD_6$: molecules with no more than 8, 7 and 6 atoms (increasing domain shift). When training on $\gD_8$, we sample 20K compounds from those with 9 atoms as our validation set and the rest for testing. This is less ideal for domain generalization evaluation since we want the validation and test set to come from different domains.

\textbf{Model Hyperparameters } For the feature extractor $\phi$, we adopt the GCN implementation from \citet{yang2019analyzing}. We use their default hyperparameters across all the datasets and baselines. Specifically, the GCN contains three convolution layers with hidden dimension 300. The predictor $f$ is a two-layer MLP with hidden dimenion 300 and ReLU activation. The model is trained with Adam optimizer for 30 epochs with batch size 50 and learning rate $\eta$ linearly annealed from $10^{-3}$ to $10^{-4}$. For RGM, we explore $\lambda \in \set{0.01,0.1}$ for each dataset. For SRGM, we explore $\lambda_g \in \set{0.1,1}$ for the classification datasets while $\lambda_g \in \set{0.01,0.1}$ for the QM9 dataset as $\lambda_g=1$ causes gradient explosion. 

\begin{figure}[t]
    \centering
    \includegraphics[width=\textwidth]{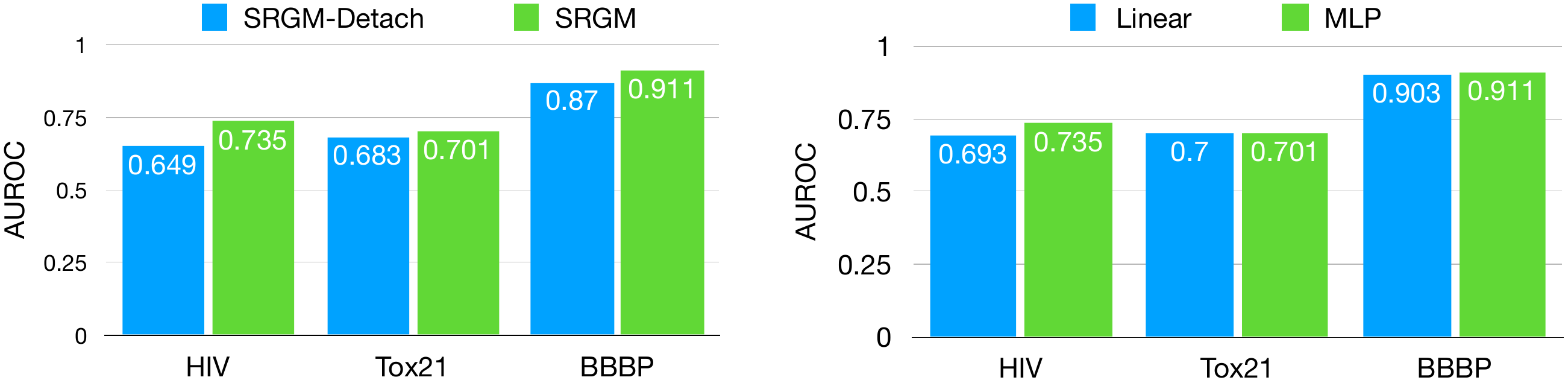}
    \caption{Ablation study of SRGM. \emph{Left}: SRGM performs better than SRGM-detach which does not update $\phi$ to optimize the scaffold classification loss $\mathcal{L}_g$. \emph{Right}: SRGM performs better than when the scaffold classifier is a MLP instead of a linear layer.}
    \label{fig:ablation_mlp}
\end{figure}

\textbf{Scaffold Classification } The scaffold classifier is trained by negative sampling since scaffolds are structured objects. Specifically, for each molecule $x_i$ in a minibatch $B$, the negative samples are the scaffolds $\set{s_k}$ of other molecules in the minibatch. The probability that $x_i$ is mapped to its correct scaffold $s_i$ is then defined as
\begin{equation}
p(s_i \;|\; x_i, B) = 
\frac{\exp\{g(\phi({x_i}))^\top g(\phi(s_i))\}}{\sum_{k\in B} \exp\{g(\phi({x_i}))^\top g(\phi(s_k))\}}
\label{eq:scaf}
\end{equation}
The scaffold classification loss is $-\sum_i \log p(s_i \;|\; x_i,B)$ for a minibatch $B$. We choose the classifier $g$ to be a two-layer MLP with hidden dimension 300 and ReLU activation. As shown in Figure~\ref{fig:ablation_mlp}, the two-layer MLP performs better than a simple linear function across multiple tasks.

\subsection{Protein Modeling}
\textbf{Data } The homology and stability dataset are downloaded from \citet{tape2019}. The size of each training environment, validation and test set are listed in Table~\ref{tab:data}.

\textbf{Model hyperparameters } For both tasks, our protein encoder is a pre-trained BERT~\citep{tape2019}. The predictor is a linear layer and the superfamily/topology classifier is a two-layer MLP whose hidden layer dimension is 768.
The model is fine-tuned with an Adam optimizer with learning rate $10^{-4}$ and linear warm up schedule. The batch size is 16 and 20 for the homology and stability task. 
For RGM and SRGM, we explore $\lambda \in \set{0.01,0.1}$ and $\lambda_g \in \set{0.1,1}$ respectively. 

\subsection{Additional Ablation Study}
In section~\ref{sec:srgm}, we mentioned that the feature extractor $\phi$ is updated to optimize the scaffold classification loss $\loss_g$. 
To study the effect of this design choice, we experiment with a variant of SRGM called SRGM-detach, in which $\phi$ is not updated to optimize the scaffold classification loss. As shown in Figure~\ref{fig:ablation_mlp}, the performance of SRGM-detach is worse than SRGM in general. This is because the scaffold classifier performs much better in SRGM and the gradient $\delta(x)$ clearly corresponds to the change of scaffold information.

\end{document}